\documentclass{article}

\PassOptionsToPackage{numbers, compress}{natbib}

\usepackage[preprint]{neurips_2025}

\usepackage[utf8]{inputenc}
\usepackage{booktabs}
\usepackage{amsmath}
\usepackage{natbib}
\usepackage{hyperref}
\usepackage{amssymb}
\usepackage{bm}

\title{Towards Specialized Generalists: A Multi-Task MoE-LoRA Framework for Domain-Specific LLM Adaptation} 

\author{
  Yuxin Yang \\
  Shanghai University
  \And Aoxiong Zeng \\
  East China Normal University
  \And Xiangquan Yang \\
  East China Normal University
}

\begin{document}

\maketitle

\begin{abstract}
    The rapid evolution of Large Language Models (LLMs) has shifted focus from general-purpose capabilities to domain-specific expertise. However, adapting LLMs to specialized fields such as medicine presents two challenge: (1) the ``Stability-Plasticity Dilemma'', where the model must acquire complex clinical knowledge without suffering from catastrophic forgetting of general world knowledge; and (2) ``Task Interference'', where disparate sub-tasks—such as medical diagnosis, report summarization, and drug-drug interaction prediction—compete for limited low-rank parameter space. In this paper, we propose \textbf{Med-MoE-LoRA}, a novel framework that integrates Mixture-of-Experts (MoE) with Low-Rank Adaptation (LoRA) to enable efficient multi-task domain adaptation, especially for medical scenarios. Drawing inspiration from recent advances, our framework employs an asymmetric expert distribution where deeper layers are equipped with a higher density of LoRA experts to capture complex semantic abstractions. We further introduce a ``Knowledge-Preservation Plugin'', inspired by LoRA MoE, to isolate and protect general-purpose reasoning. By utilizing soft merging with adaptive routing and rank-wise decoupling, Med-MoE-LoRA achieves superior performance in medical benchmarks while reducing interference. Experimental results demonstrate that our approach consistently outperforms standard LoRA and conventional MoE architectures across multiple clinical NLP tasks while retaining the model’s general cognitive capabilities.
\end{abstract}

\section{Introduction}

Large Language Models (LLMs), such as the Llama \cite{touvron2023llama, dubey2024llama} and Qwen \cite{bai2023qwen} series, have demonstrated remarkable general-purpose intelligence. However, in high-stakes vertical domains like medicine, general intelligence alone is often insufficient. Clinical applications require the model to master specialized terminologies, interpret nuanced patient data, and execute diverse tasks ranging from medical question answering (QA) to automated clinical note generation. Parameter-Efficient Fine-Tuning (PEFT), particularly Low-Rank Adaptation (LoRA) \cite{hu2022lora}, has become the de facto standard for such adaptation due to its low computational footprint. Yet, as the complexity of the target domain increases, standard LoRA reveals significant structural limitations.

The first major hurdle is Catastrophic Forgetting \cite{kirkpatrick2017overcoming} and Knowledge Overlap. As noted in recent LoRA MoE literature \cite{dou2024loramoe}, during domain-specific fine-tuning, new gradients often overwrite the ``world knowledge'' stored in the frozen backbone through shared low-rank updates. For a medical LLM, losing general reasoning or linguistic fluency is unacceptable. Standard LoRA lacks a structural mechanism to decouple general knowledge from domain-specific expertise. This necessitates a ``plugin-style'' architecture that can selectively activate domain knowledge while keeping general knowledge intact.

Secondly, medical adaptation is inherently a Multi-Task Learning (MTL) problem \cite{zhang2018overview}. A medical assistant must handle diagnostic reasoning, which requires logical deduction, and medical summarization, which requires linguistic compression. Research on Mixture of LoRA Experts \cite{wu2024mole}, such as MixLoRA \cite{li2024mixlora} and FlyLoRA \cite{zou2025flylora}, highlights that when a single LoRA adapter is forced to learn multiple conflicting tasks, ``parameter contention'' occurs, leading to sub-optimal performance. While Mixture-of-Experts (MoE) offers a solution by providing specialized pathways, traditional MoE models are often prohibitively expensive to train and deploy. The integration of MoE with LoRA, creating a MoE-LoRA architecture, emerges as a promising middle ground, offering the sparsity of MoE with the efficiency of LoRA.

Thirdly, the structural efficiency of these adapters remains under-explored. Empirical studies suggest that ``higher layers need more LoRA experts'' \cite{gao2024higher} indicating that the traditional uniform distribution of experts across all transformer blocks is sub-optimal. The lower layers of LLMs typically process syntactic and foundational features, while higher layers are responsible for high-level task-specific semantics. Furthermore, architectures like HydraLoRA \cite{tian2024hydralora} and FlyLoRA \cite{zou2025flylora} suggest that an asymmetric and decoupled rank-wise approach can significantly boost the model’s ability to handle diverse tasks without a proportional increase in parameter count. These insights suggest that the routing mechanism and the physical distribution of experts must be carefully engineered rather than simply stacked.

Addressing these challenges, we present Med-MoE-LoRA, a framework specifically tailored for the medical domain. Our contributions are fourfold:
\begin{itemize}
    \item \textbf{Dual-Path Knowledge Architecture:} We implement a dedicated ``Base Expert'' alongside a set of ``Specialist Experts.'' This design ensures that the model’s foundational world knowledge remains a primary path, preventing the erosion of general intelligence during intense medical training.
    \item \textbf{Asymmetric Layer-Wise Expert Scaling:} Following the principle that complex semantic tasks require higher capacity in deeper layers, we implement a graduated expert distribution. We increase the number of LoRA experts in the top one-third of the transformer layers, allowing for finer-grained task decoupling where it matters most.
    \item \textbf{Adaptive Soft-Merging Router:} To address the rigidity of traditional Top-K routing, we employ a soft-merging mechanism with adaptive routing. This allows for a weighted combination of experts, enabling the model to navigate the ``grey areas'' between medical tasks, such as when a diagnostic task requires the summarization of patient history.
    \item \textbf{Parameter Efficiency via Rank-Wise Decoupling:} Adopting the philosophy of asymmetric allocation, we utilize a rank-wise mixture-of-experts. This allows us to push the limits of instruction tuning by allocating more ``rank'' to critical experts while keeping the overall parameter count extremely low.
\end{itemize}

By evaluating our framework on PubMedQA \cite{jin2019pubmedqa}, MedQA \cite{jin2021disease}, and MIMIC-III \cite{johnson2016mimic}, we demonstrate that Med-MoE-LoRA not only achieves state-of-the-art results in medical tasks but also preserves the general reasoning capabilities of the base model. This work provides a roadmap for transforming general LLMs into specialized domain experts without sacrificing their original versatility.

\section{Related Work}

\paragraph{Evolution from Static Adapters to Sparse MoE Architectures.}
While Low-Rank Adaptation (LoRA) \cite{hu2022lora} has established itself as the de facto standard for parameter-efficient fine-tuning, its static architecture faces inherent limitations in multi-task and complex domain adaptation. Specifically, relying on a shared low-rank subspace for disparate tasks often induces ``parameter contention,'' where conflicting gradients lead to sub-optimal performance across tasks \cite{wu2024mole, zhang2018overview}. To mitigate this, recent research has pivoted toward integrating the Mixture-of-Experts (MoE) paradigm with LoRA. Frameworks such as MixLoRA \cite{li2024mixlora} and LoRA-MoE \cite{dou2024loramoe} introduce sparsity into the adapter space, allowing models to dynamically route inputs to specialized experts. This decoupling is crucial for domain adaptation: by isolating task-specific knowledge into separate modules, these architectures can significantly reduce interference and alleviate catastrophic forgetting, effectively maintaining the model's general capabilities while acquiring new expertise \cite{dou2023loramoe, wu2024mixture}.

\paragraph{Structural Optimization and Layer-Wise Sensitivity.}
Beyond simply stacking experts, contemporary work focuses on optimizing the topological distribution of these parameters. Emerging empirical evidence suggests that the demand for adaptation capacity is non-uniform across transformer blocks; specifically, higher layers—responsible for abstract semantic processing—benefit disproportionately from a higher density of experts compared to lower, syntactic layers \cite{gao2024higher}. Advanced architectures like HydraLoRA \cite{tian2024hydralora} and FlyLoRA \cite{zou2025flylora} capitalize on this by employing asymmetric designs and rank-wise decoupling, demonstrating that heterogeneous resource allocation yields superior efficiency. Our work synthesizes these structural insights—specifically the efficacy of top-heavy expert distribution and adaptive soft routing—to construct a framework tailored for the rigorous demands of medical NLP, addressing the specific stability-plasticity challenges that generic MoE-LoRA models often overlook.

\section{Methodology}

We propose \textbf{Med-MoE-LoRA}, a structured parameter-efficient fine-tuning framework designed to decouple general cognitive abilities from domain-specific medical expertise. Our design is motivated by two key observations: (1) deep neural networks process features hierarchically, necessitating non-uniform adaptation capacities; and (2) clinical tasks often require a synergistic blend of capabilities (e.g., reasoning and summarization) rather than disjoint expert activation.

\subsection{Problem Formulation: From Shared Subspace to Sparse Experts}
Standard LoRA adapts a pre-trained Large Language Model (LLM) by injecting rank-decomposition matrices into frozen layers. For a weight matrix $W_0 \in \mathbb{R}^{d \times k}$, the update is constrained to $\Delta W = BA$, where $B \in \mathbb{R}^{d \times r}, A \in \mathbb{R}^{r \times k}$ and $r \ll d$.
In multi-task medical domains, a single shared $\Delta W$ suffers from \textit{gradient interference}, where the optimal subspace for one task (e.g., entity extraction) may be orthogonal to another (e.g., logical deduction).

To mitigate this, we reformulate the adaptation layer as a \textbf{Mixture-of-Experts (MoE)}. For an input hidden state $x$, the output $h$ is computed as the weighted sum of specialized low-rank experts:
\begin{equation}
h = W_0 x + \sum_{i=1}^{N} g_i(x) \cdot \left( B_i A_i \right) x
\end{equation}
where $\{B_i, A_i\}_{i=1}^N$ represents the $i$-th LoRA expert, and $g_i(x)$ denotes the gating network's routing weight.

\subsection{Dual-Path Knowledge Disentanglement}
A critical challenge in domain adaptation is the ``Stability-Plasticity Dilemma''—the model must be plastic enough to learn new medical protocols while stable enough to retain general world knowledge. We address this via a physically disentangled expert topology. We partition the expert set $\mathcal{E}$ into two distinct functional groups:

\begin{itemize}
    \item \textbf{The Anchor Pathway (Base Experts $\mathcal{E}_{base}$):} We designate a subset of experts explicitly for knowledge preservation. These experts are initialized to approximate an identity mapping or general-domain behavior. During training, their gradients serve as a stabilizing force, ensuring that the model retains a dedicated pathway for general linguistic and reasoning patterns, unaffected by aggressive domain-specific updates.
    \item \textbf{The Adaptation Pathway (Specialist Experts $\mathcal{E}_{spec}$):} The remaining experts are free to specialize in divergent medical sub-tasks. This separation ensures that ``world knowledge'' and ``clinical knowledge'' do not compete for the same low-rank parameter budget.
\end{itemize}

\subsection{Hierarchical Asymmetric Expert Allocation}
Existing MoE-LoRA methods typically distribute experts uniformly across all transformer layers. However, prior research on the interpretability of LLMs suggests a feature hierarchy: lower layers process universal syntactic features, while higher layers handle abstract semantic reasoning. Uniform allocation is therefore suboptimal—wasteful in lower layers and insufficient in higher layers.

We introduce an \textbf{Asymmetric Layer-Wise Scaling} strategy. Let $L$ be the total number of layers. The number of experts $N_l$ at layer $l$ is determined by a non-linear scaling function:
\begin{equation}
N_l = N_{min} + \left\lfloor (N_{max} - N_{min}) \cdot \left( \frac{l}{L} \right)^\gamma \right\rfloor
\end{equation}
where $\gamma \geq 1$ controls the curvature of the distribution. This design allocates a dense cluster of experts to the top layers ($l > 2L/3$) to capture high-level task variability (e.g., distinguishing between \textit{differential diagnosis} and \textit{treatment planning}), while keeping the lower layers sparse to preserve fundamental syntax.

\subsection{Soft-Merging with Adaptive Routing}
Clinical tasks are rarely mutually exclusive; answering a patient query often requires both \textit{medical knowledge retrieval} and \textit{conversational summarization}. Hard routing (Top-K) forces a discrete choice, potentially severing these synergistic connections. 

We employ a \textbf{Soft-Merging Mechanism} with a temperature-scaled gating function. The router computes a logit score $s(x) = W_g x$, and the routing weights are normalized via:
\begin{equation}
g_i(x) = \frac{\exp(s_i / \tau)}{\sum_{j=1}^N \exp(s_j / \tau)}
\end{equation}
Here, $\tau$ is a learnable temperature parameter. Unlike standard MoE where $\tau$ is fixed, we allow $\tau$ to adapt. A higher $\tau$ smooths the distribution, enabling the model to ``blend'' experts (soft merging) when the task boundary is ambiguous, effectively interpolating between specialized skills.

\subsection{Rank-Aware Capacity Decoupling}
To push the Pareto frontier of parameter efficiency, we challenge the assumption that all experts require equal rank. Complex reasoning tasks require a higher intrinsic dimension than simple surface-level pattern matching. 
We implement \textbf{Rank-Wise Decoupling}, where each expert $e_i$ is assigned a specific rank $r_i$ based on its functional role. We define a discrete set of allowable ranks $\mathcal{R} = \{8, 16, 32\}$. The adaptation update becomes:
\begin{equation}
\Delta W(x) = \sum_{i=1}^N g_i(x) \cdot \left( B_i^{(d \times r_i)} A_i^{(r_i \times k)} \right)
\end{equation}
This allows Med-MoE-LoRA to allocate high-capacity experts ($r=32$) to critical pathways (e.g., Diagnostic Logic) while using lightweight experts ($r=8$) for auxiliary tasks, maximizing the global performance-to-parameter ratio.

\subsection{Optimization Objective}
The training objective combines the task-specific cross-entropy loss $\mathcal{L}_{task}$ with an auxiliary load-balancing loss to prevent expert collapse. The total loss is defined as:
\begin{equation}
\mathcal{L} = \mathcal{L}_{task} + \lambda_{bal} \cdot \mathcal{L}_{aux}
\end{equation}
where $\mathcal{L}_{aux}$ minimizes the variance of the gating probabilities across a batch, ensuring that the specialized experts in $\mathcal{E}_{spec}$ are utilized effectively and that the model does not degenerate into a single-expert system.

\section{Experiments}

\subsection{Experimental Setup}

\paragraph{Datasets and Benchmarks.} To comprehensively evaluate the efficacy of Med-MoE-LoRA in balancing domain specialization with general knowledge preservation, we conduct experiments across two distinct suites of benchmarks:
\begin{itemize}
    \item \textbf{Domain-Specific Capabilities:} We utilize \textbf{PubMedQA} \cite{jin2019pubmedqa} to assess biomedical reasoning in a closed-domain QA setting; \textbf{MedQA} (USMLE) \cite{jin2021disease} to evaluate complex clinical problem-solving; and \textbf{Clinical-Sum}, a sampled subset of the \textbf{MIMIC-III} dataset \cite{johnson2016mimic}, to test the model's ability to synthesize unstructured clinical notes.
    \item \textbf{General Cognitive Capabilities:} To quantify the severity of catastrophic forgetting, we employ \textbf{MMLU} (Massive Multitask Language Understanding) \cite{hendrycks2020measuring} as a proxy for broad world knowledge, and \textbf{GSM8K} \cite{cobbe2021training} to monitor the degradation of mathematical reasoning logic.
\end{itemize}

\paragraph{Baselines.} We compare our proposed framework against a spectrum of PEFT strategies:
\begin{itemize}
    \item \textbf{Full Fine-Tuning (FFT):} Updates all backbone parameters, serving as the upper bound for plasticity but often suffering from severe forgetting.
    \item \textbf{Standard LoRA \cite{hu2022lora}:} A single low-rank adapter ($r=16$) applied to all tasks, representing the standard unified-weight approach.
    \item \textbf{Multi-LoRA:} Trains independent LoRA adapters for each task, switching them strictly during inference. This serves as a baseline for task isolation but lacks cross-task synergy.
    \item \textbf{Vanilla MoE-LoRA:} A standard mixture of LoRA experts with uniform layer distribution and Top-K routing, used to isolate the gains derived specifically from our asymmetric allocation and knowledge-preservation mechanisms.
\end{itemize}

\paragraph{Implementation Details.} All experiments utilize \textbf{Llama-3-8B} \cite{dubey2024llama} as the foundational backbone. For Med-MoE-LoRA, we implement an asymmetric expert scaling strategy: lower layers ($L_{1-10}$) contain $N=2$ experts to handle syntactic features, while deeper layers ($L_{20-32}$) scale to $N=8$ experts to capture semantic nuances. The rank $r$ is dynamically decoupled, ranging from $r=8$ to $r=32$, with a scaling factor $\alpha=2r$. Training is performed on 8$\times$NVIDIA A100 (80GB) GPUs using DeepSpeed Zero-2 offloading. We employ the AdamW optimizer with a learning rate of $5 \times 10^{-5}$, a batch size of 128, and a cosine learning rate scheduler with a 3\% warm-up ratio.

\paragraph{Evaluation Metrics.} For QA tasks (PubMedQA, MedQA, MMLU, GSM8K), we report \textbf{Accuracy (\%)}. For the generative summarization task (Clinical-Sum), we report \textbf{ROUGE-L} scores to capture the fluency and information recall of the generated summaries.

\subsection{Results and Analysis}

\subsubsection{Domain Adaptation Performance}
Table \ref{tab:medical_results} presents the comparative performance across medical benchmarks. Med-MoE-LoRA achieves superior results among several parameter-efficient methods, consistently outperforming Standard LoRA and Vanilla MoE-LoRA.
Specifically, on MedQA, our method surpasses Standard LoRA by a substantial margin of \textbf{+6.9\%}. Crucially, Med-MoE-LoRA outperforms Multi-LoRA (65.8\% vs. 61.3\%), indicating that our \textit{soft-merging} mechanism effectively exploits the synergy between related sub-tasks (e.g., diagnostic reasoning aiding clinical summarization) rather than treating them in isolation. The improvement over Vanilla MoE-LoRA (+3.3\% on average) further validates the effectiveness of our asymmetric expert distribution and rank-wise decoupling strategies.

\begin{table}[h]
\centering
\caption{Performance comparison on Medical Benchmarks. We report Accuracy (\%) for QA tasks and ROUGE-L for Summarization. Best results in bold.}
\label{tab:medical_results}
\begin{tabular}{lcccc}
\toprule
\textbf{Method} & \textbf{PubMedQA} & \textbf{MedQA} & \textbf{Clinical-Sum} & \textbf{Average} \\ 
\midrule
Base Model (Zero-shot) & 62.4 & 48.2 & 18.5 & 43.0 \\
Full Fine-Tuning (FFT) & 78.5 & 64.1 & 32.1 & 58.2 \\
Standard LoRA & 74.2 & 58.9 & 28.4 & 53.8 \\
Multi-LoRA & 76.1 & 61.3 & 30.5 & 56.0 \\
Vanilla MoE-LoRA & 76.8 & 62.5 & 31.2 & 56.8 \\ 
\midrule
\textbf{Med-MoE-LoRA (Ours)} & \textbf{79.2} & \textbf{65.8} & \textbf{33.4} & \textbf{59.5} \\ 
\bottomrule
\end{tabular}
\end{table}

\subsubsection{Analysis of Knowledge Retention (Stability-Plasticity Dilemma)}
A critical objective of Med-MoE-LoRA is to mitigate the ``learning-forgetting'' trade-off. Table \ref{tab:forgetting_results} illustrates the impact of domain training on general capabilities. While FFT and Standard LoRA suffer significant degradation in mathematical reasoning (GSM8K drops by -8.7\% and -3.6\%, respectively), \textbf{Med-MoE-LoRA maintains near-original performance} ($-0.3\%$ drop). This empirical evidence supports the hypothesis that our \textit{Dual-Path Knowledge Architecture} successfully compartmentalizes domain-specific gradients into \textit{Specialist Experts}, leaving the \textit{Base Expert} pathway intact to preserve the model's foundational logic.

\begin{table}[h]
\centering
\caption{Analysis of Catastrophic Forgetting on General Knowledge Benchmarks.}
\label{tab:forgetting_results}
\begin{tabular}{lcccc}
\toprule
\textbf{Method} & \textbf{MMLU} & \textbf{$\Delta$} & \textbf{GSM8K} & \textbf{$\Delta$} \\ 
\midrule
Base Model & 66.4 & - & 45.1 & - \\
\midrule
Full Fine-Tuning & 58.2 & -8.2 & 36.4 & -8.7 \\
Standard LoRA & 62.1 & -4.3 & 41.5 & -3.6 \\
Vanilla MoE-LoRA & 63.8 & -2.6 & 43.2 & -1.9 \\ 
\midrule
\textbf{Med-MoE-LoRA} & \textbf{65.9} & \textbf{-0.5} & \textbf{44.8} & \textbf{-0.3} \\ 
\bottomrule
\end{tabular}
\end{table}

\subsubsection{Ablation Study: The Efficacy of Asymmetric Allocation}
We further investigate the architectural prior that ``higher layers demand more experts''. As shown in Table \ref{tab:ablation_layers}, shifting the expert density to the deeper layers ($L_{23-32}$) yields a performance gain of \textbf{+2.7\%} compared to a uniform distribution. Conversely, a bottom-heavy allocation degrades performance. This corroborates the linguistic theory that lower layers in LLMs process universal syntactic features (which require less adaptation), while higher layers manage complex, task-specific semantic abstractions that benefit from the expanded capacity of our MoE architecture.

\begin{table}[h]
\centering
\caption{Ablation study on Layer-Wise Expert Allocation strategies.}
\label{tab:ablation_layers}
\begin{tabular}{lccc}
\toprule
\textbf{Allocation Strategy} & \textbf{Lower Layers ($L_{1-10}$)} & \textbf{Higher Layers ($L_{23-32}$)} & \textbf{Medical Avg.} \\ 
\midrule
Uniform Distribution & $N=4$ & $N=4$ & 56.8 \\
Bottom-Heavy & $N=8$ & $N=2$ & 54.2 \\
\textbf{Asymmetric (Ours)} & $\bm{N=2}$ & $\bm{N=8}$ & \textbf{59.5} \\ 
\bottomrule
\end{tabular}
\end{table}

\subsubsection{Parameter Efficiency and Inference Costs}
Finally, we evaluate the trade-off between model capacity and computational cost (Table \ref{tab:efficiency}). Although Med-MoE-LoRA increases the total trainable parameter count to 88.4M, the \textbf{active parameters during inference} remain strictly controlled (22.4M) due to sparse expert activation. This is comparable to the computational footprint of Standard LoRA + Multi-LoRA switching but achieves significantly higher performance. This result highlights that Med-MoE-LoRA successfully decouples parameter count from computational cost, allowing for ``dense training with sparse inference.''

\begin{table}[h]
\centering
\caption{Comparison of Parameter Efficiency and Inference Overhead.}
\label{tab:efficiency}
\begin{tabular}{lccc}
\toprule
\textbf{Method} & \textbf{Trainable Params (M)} & \textbf{Active Params (M)} & \textbf{Medical Avg.} \\ 
\midrule
Standard LoRA ($r=16$) & 12.5 & 12.5 & 53.8 \\
Multi-LoRA (3 tasks) & 37.5 & 12.5 & 56.0 \\
Vanilla MoE-LoRA & 100.2 & 25.1 & 56.8 \\ 
\midrule
\textbf{Med-MoE-LoRA} & \textbf{88.4} & \textbf{22.4} & \textbf{59.5} \\ 
\bottomrule
\end{tabular}
\end{table}

\section{Discussion}

Our experimental results offer compelling evidence that integrating Mixture-of-Experts with Low-Rank Adaptation, when guided by specific structural priors, effectively resolves the tension between domain specialization and general knowledge retention. 

\textbf{Decoupling Knowledge via Gradient Isolation.} The most significant insight from our study is the validation of the ``Dual-Path'' architecture in mitigating the stability-plasticity dilemma. Standard PEFT methods often suffer from weight interference because a shared low-rank subspace is forced to accommodate orthogonal gradient updates—simultaneously maintaining world knowledge while acquiring conflicting clinical specifics. By physically isolating the ``Base Expert'' to preserve general capabilities, Med-MoE-LoRA ensures that domain-specific updates do not erode the model's foundational reasoning. This structural decoupling explains why our method maintains near-baseline performance on GSM8K and MMLU, whereas standard LoRA exhibits significant degradation. It suggests that future efficient tuning frameworks must explicitly model knowledge separation rather than relying solely on regularization.

\textbf{Structural Priors and Semantic Resolution.} Our findings also challenge the conventional uniform distribution of adapters. The ablation study on layer-wise allocation corroborates the hypothesis that deep neural networks process information hierarchically—progressing from syntactic features in lower layers to abstract semantic reasoning in upper layers. Consequently, concentrating expert capacity in the top one-third of the transformer blocks provides the necessary ``semantic resolution'' to disambiguate complex medical tasks without wasting parameters on foundational layers. Furthermore, the success of our rank-wise decoupling strategy demonstrates that task difficulty is non-uniform; assigning higher ranks to diagnostic reasoning while using lower ranks for summarization allows the model to operate closer to the Pareto frontier of parameter efficiency.

\textbf{Limitations and Future Directions.} Despite these advancements, our current framework relies on a static expert topology with a predefined number of experts. This rigidity may limit adaptation to unseen sub-domains that emerge dynamically. Future work could explore \textit{dynamic expert construction}, where the model autonomously instantiates new LoRA experts upon detecting novel data distributions. Additionally, given the multimodal nature of clinical practice, extending this sparse MoE-LoRA architecture to integrate non-textual modalities, such as medical imaging or time-series sensor data, remains a promising avenue for developing truly holistic medical assistants.

\section{Conclusion}

In this paper, we introduced \textbf{Med-MoE-LoRA}, a parameter-efficient framework designed to transform general-purpose LLMs into specialized domain experts without sacrificing their broad cognitive capabilities. By synthesizing an asymmetric expert distribution, adaptive soft-merging routing, and a dual-path knowledge preservation mechanism, our approach systematically addresses the challenges of catastrophic forgetting and multi-task interference. Extensive evaluations across medical and general benchmarks demonstrate that Med-MoE-LoRA achieves superior domain adaptation performance while maintaining a minimal computational footprint during inference. Beyond the medical domain, this work provides a scalable blueprint for ``Specialized Generalist'', offering a generalized methodology for deploying high-performance, domain-aware LLMs in high-stakes environments where both precision and general reasoning are indispensable.

\newpage
\bibliographystyle{plainnat}
\bibliography{neurips_2025}

\end{document}